\pdfoutput=1

\documentclass[11pt]{article}

\usepackage[preprint]{acl}
\usepackage[most]{tcolorbox}
\usepackage{times}
\usepackage{multirow}
\usepackage{latexsym}
\usepackage{amsmath}
\usepackage[T1]{fontenc}
\usepackage{booktabs}
\usepackage{threeparttable}
\usepackage[utf8]{inputenc}

\usepackage{booktabs}
\usepackage{pifont}
\usepackage{array}
\newcommand{\cmark}{\ding{51}}
\newcommand{\xmark}{\ding{55}}
\usepackage{inconsolata}

\usepackage{graphicx}
\usepackage{enumitem}
\title{MedCalc-Pro: Solving Complex Medical Calculations with LLM Agents}

\author{
\textbf{Siran Zhao\textsuperscript{\rm $\diamondsuit$}\thanks{~~Equal Contribution.}},
\textbf{Ruihui Hou\textsuperscript{\rm $\diamondsuit$}\footnotemark[1]},
\textbf{Ziyue Huai\textsuperscript{\rm $\diamondsuit$}},
\textbf{Chennuo Zhang\textsuperscript{\rm $\diamondsuit$}},
\\
\textbf{Tong Ruan\textsuperscript{\rm $\diamondsuit$}\thanks{Corresponding authors}}
\\
\textsuperscript{\rm $\diamondsuit$}East China University of Science and Technology, Shanghai, China,
\\
}

\begin{document}
\maketitle
\begin{abstract}
Current benchmarks for evaluating large language models (LLMs) in medical calculation are largely based on simplified settings, where each patient case corresponds to a single calculator and the required tool is explicitly specified in the query. However, real clinical scenarios often require multiple calculators for joint evaluation, nested-scale calculation, and fuzzy queries that do not directly specify the target calculator. To this end, we propose a new medical calculation benchmark, MedCalc-Pro, which covers three progressively challenging task settings: single-calculator, multi-calculator, and nested-calculator calculation settings. MedCalc-Pro contains 2,268 real-world clinical cases, covering 77 medical calculators across 14 clinical departments. Meanwhile, to address the limited performance of existing frameworks and methods in complex clinical scenarios, we further propose a more generalizable agent framework that supports multi-tool selection and nested-tool calling, while suppressing parameter error propagation through structured validation and evidence review. We conduct systematic comparisons across open-source, closed-source, and medical-specialized LLMs, and the results show that our framework achieves the best performance across all three task settings. This work provides a new benchmark and method for evaluating and applying LLMs in challenging medical calculation scenarios.

\end{abstract}

\section{Introduction}

\begin{figure}[t]
\centering
\includegraphics[width=\linewidth]{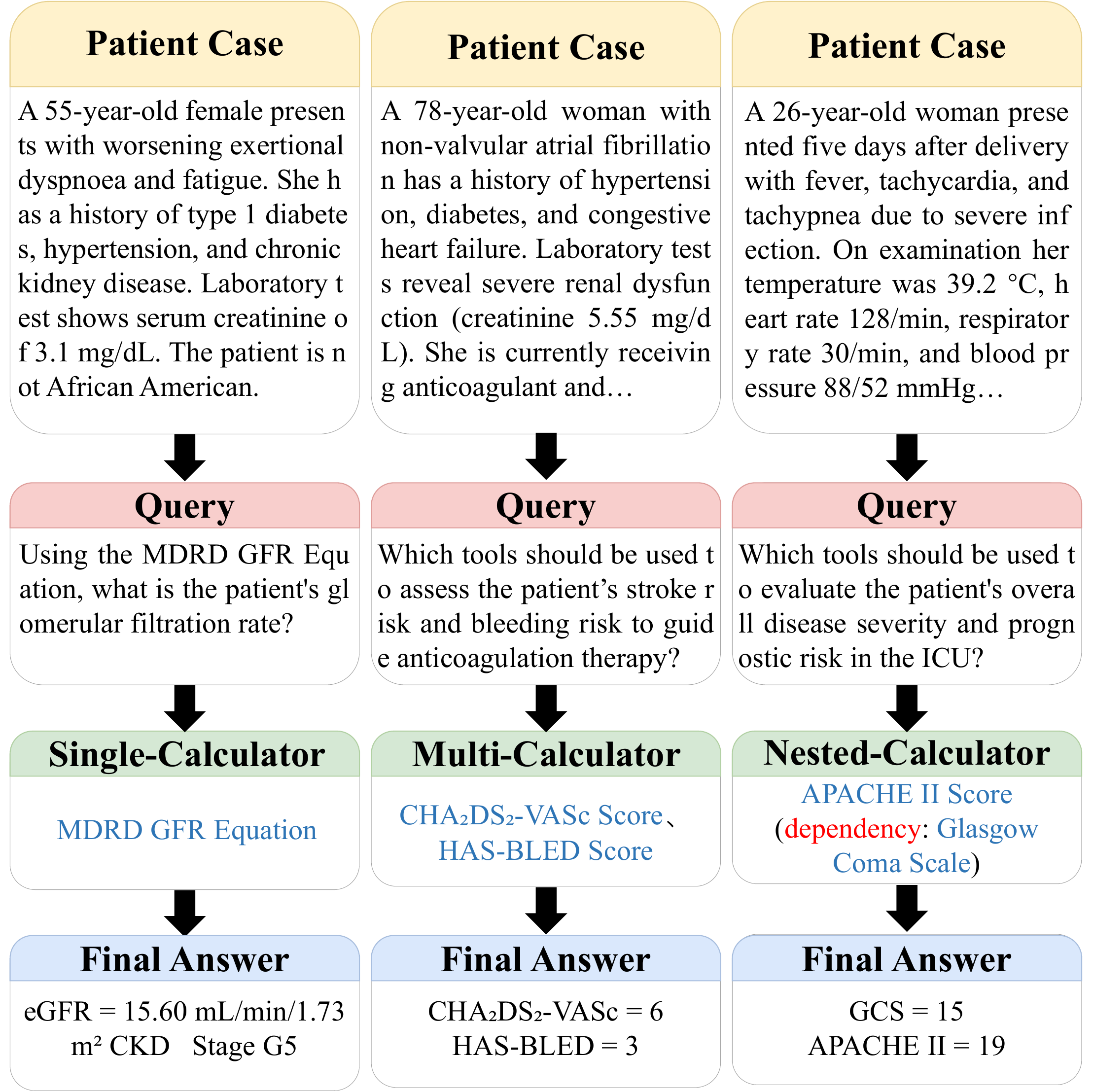}
\caption{Illustrative examples of three medical calculation scenarios:
single-calculator, multi-calculator, and nested-calculator scenarios.}
\label{intro_example}
\end{figure}

Large language models (LLMs) have recently achieved remarkable progress in information understanding~\citep{large} and complex reasoning~\citep{training}. However, they still exhibit clear limitations on certain tasks that precise execution by external tools, such as numerical computation and rule-based clinical score calculation~\citep{metatool,clinicalprediction}. This issue is particularly salient in the medical domain. In clinical practice, medical calculators are widely used for risk assessment and prognosis prediction~\citep{prediction}. 
Consequently, there has been growing interest in integrating LLMs with medical calculators.

However, existing medical calculation benchmarks, including MedCalc-Bench~\citep{medcalc}, CMedCalc-Bench~\citep{cmedcalc}, OpenMedCalc~\citep{augmentation}, CalcQA~\citep{menti}, and AgentMD~\citep{agentmd}, remain largely restricted to simplified settings. These benchmarks typically pair an explicit target-calculator query with a clinical case description, with each case corresponding to only one calculator. Such settings are insufficient for capturing more realistic medical calculation scenarios. In practice, multiple calculators often need to be used jointly, and some complex calculators further depend on sub-calculator outputs, forming nested tool-calling chains.

\begin{table*}[t]
\small
\resizebox{\textwidth}{!}{%
\begin{tabular}{lccccc}
\toprule
\textbf{Benchmarks} & \textbf{Fuzzy Query} & \textbf{Proc. Eval.} & \textbf{Multi-calculator} & \textbf{Nested-calculator} & \textbf{Size} \\
\midrule
RiskQA~\citep{agentmd}                & \xmark & \cmark & \xmark & \xmark & 350 \\
MedRaC~\citep{medrac}                & \xmark & \cmark & \xmark & \xmark & 1,048 \\
CMedCalc-Bench~\citep{cmedcalc}      & \xmark & \cmark & \xmark & \xmark & 1,143 \\
OpenMedCalc~\citep{augmentation}     & \xmark & \xmark & \xmark & \xmark & 287 \\
MedCalc-Bench~\citep{medcalc}        & \xmark & \xmark & \xmark & \xmark & 1,047 \\
CalcQA~\citep{menti}                 & \cmark & \cmark & \xmark & \xmark & 100 \\
MedCalc-Pro (Ours)                                & \cmark & \cmark & \cmark & \cmark & 2,268 \\
\bottomrule
\end{tabular}%
}
\caption{Comparison of medical calculator benchmarks. 
``Fuzzy Query'' indicates whether the benchmark includes goal-driven queries that do not explicitly specify the target calculator. 
``Proc. Eval.'' denotes whether process-level evaluation is provided. 
``Multi-calculator'' indicates whether cases may require multiple calculators, 
and ``Nested-calculator'' indicates whether cases involve dependencies between calculators.}
\label{tab:benchmark_comparison}

\end{table*}

To address these limitations, we propose a new medical calculation benchmark, \textbf{MedCalc-Pro}, for systematically evaluating LLMs in complex medical calculation scenarios. The benchmark covers three progressively challenging task settings: single-calculator, multi-calculator, and nested-calculator calculation. Specifically, single-calculator task corresponds to the basic setting where a single patient case requires one calculator; multi-calculator task corresponds to tasks where a single case requires multiple calculators for joint evaluation; and nested-calculator task further requires the model to identify dependencies among calculators. Figure~\ref{intro_example} illustrates these three task settings. 
MedCalc-Pro contains 2,268 real-world clinical cases, covering 77 medical calculators across 14 clinical departments. 
Compared with existing benchmarks, MedCalc-Pro not only covers task settings that are closer to real clinical usage but also places greater emphasis on clinically goal-driven queries, thereby avoiding direct exposure of the tool name to the model and imposing stricter requirements on query understanding, tool selection, parameter extraction, and multi-step execution.
Table~\ref{tab:benchmark_comparison} summarizes the key differences between existing medical calculator benchmarks and MedCalc-Pro.

Meanwhile, to address the limited generalization of existing methods and frameworks in complex settings, we propose a more generalizable medical calculator agent framework. The framework follows realistic medical calculation workflows and consists of four stages: Query Rewriting, Retrieval and Reranking, Tool Selection, and Tool Execution. Query Rewriting enhances clinical-intent representation through multi-dimensional rewriting of the original query. Retrieval and Reranking identifies relevant candidate calculators from the tool repository. 
Tool Selection enables the model to choose the appropriate calculators for the clinical task. 
Finally, Tool Execution performs dependency-aware nested-calculator invocation, while structured validation and evidence auditing are introduced to suppress error propagation and improve the robustness of the final calculation.

Our main contributions are as follows:
\begin{itemize} 
    \item We construct a new medical calculation benchmark, MedCalc-Pro, to enable more realistic evaluation of LLMs across single-calculator, multi-calculator, and nested-calculator scenarios.
    \item We propose a generalizable agent framework for complex medical calculation, supporting multi-calculator selection and nested-calculator execution through query rewriting, retrieval and reranking, tool selection, and dependency-aware tool execution with structured validation and evidence review.
    \item We compare our method with representative frameworks across open-source, closed-source, and medical-specialized models. Results show that it achieves the best performance across all three task settings and demonstrates strong robustness and generalization in complex scenarios.
\end{itemize}

\section{Related Works}
\subsection{Medical Calculator Benchmarks}
In recent years, as LLMs have been applied to an increasingly wide range of real-world tasks~\citep{agi,gptmedical}, their performance in domain-specific applications has attracted growing attention~\citep{med-palm}. In the medical domain, many early datasets, such as MedQA~\citep{medQA}, PubMedQA~\citep{pubmedqa}, MedMCQA~\citep{medmcqa}, and MMLU medical subjects~\citep{MMLU}, primarily adopt multiple-choice and open-ended formats, which assess whether LLMs can recall and reason over medical knowledge. 
Furthermore, several recent studies have proposed benchmarks specifically targeting medical calculation tasks such as CMedCalc-Bench, CalcQA, and MedCalc-Bench. Despite these efforts, existing benchmarks typically assume a single-calculator setting, and the task description often explicitly specifies which calculator should be used. In real-world clinical scenarios, however, a single patient case often requires evaluation using multiple medical calculators, and clinicians may not explicitly indicate which calculators should be applied. To better reflect realistic clinical workflows, we constructed the MedCalc-Pro that supports three types of  settings: single-calculator,  multi-calculator, and nested-calculator calculation.

\subsection{LLM-based Methods for Medical Calculator Execution}
Relying solely on the model for the entire computation process poses a significant limitation. The model's numerical reasoning and calculation performance is often subpar~\citep{calculate}, reducing transparency and making intermediate steps difficult to verify, which is particularly critical in medical scenarios. To improve reliability and transparency, more recent studies have explored integrating LLMs with external tools~\citep{toolformer}, such as code interpreters~\citep{code,program} and external APIs~\citep{api}.
This approach has also been widely adopted in medical calculation tasks, such as MeNTi~\citep{menti}, and AgentMD enable LLMs to interact with structured medical calculators, allowing deterministic execution of clinical scoring programs. 
However, these frameworks also typically assume that each patient case corresponds to a single-calculator calculation, limiting their applicability in realistic clinical scenarios. 
These challenges motivate our framework design, which supports flexible multi-calculator and nested-calculator execution through a structured agent workflow.
\section{Methodology}
\subsection{Task Definition}
Given a patient's electronic medical record (EMR) and a query $Q$, the task aims to select appropriate medical calculators from a set of candidates and compute the corresponding answer based on the patient's clinical information. 
As illustrated in Figure~\ref{intro_example}, when the query is ``Which tools should be used to assess the patient’s stroke risk and bleeding risk to guide anticoagulation therapy?'', the model first selects the relevant calculators, CHA$_2$DS$_2$-VASc Score and HAS-BLED Score. 
It then applies the scoring rules of these calculators using the patient's clinical information and finally produces the corresponding scores of 6 and 3.
\begin{figure*}[t]
    \centering
    \includegraphics[width=\textwidth]{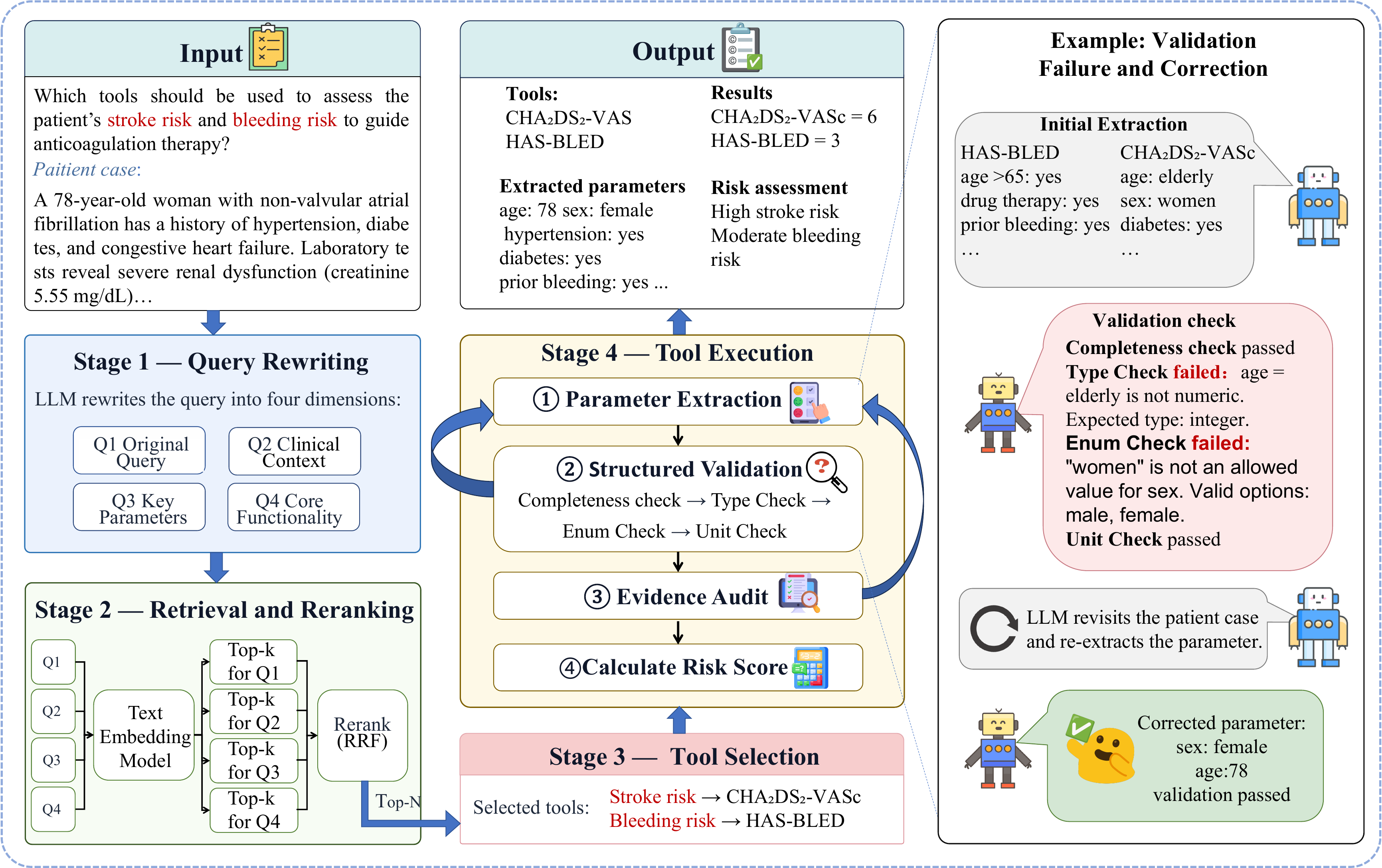}
    \caption{Overview of our framework. This agent contains four stages: Query Rewriting, Retrieval and Reranking, Tool Selection, and Tool Execution.}
    \label{fig:your_label}
\end{figure*}

\subsection{Benchmark Creation}
To evaluate the capability of LLMs in medical calculation tasks, 
we construct MedCalc-Pro, a benchmark covering multiple evaluation scenarios. 
The benchmark includes three types of calculation tasks: single-calculator, 
multi-calculator, and nested-calculator.


For the \textbf{single-calculator} task, we use MedRaC as the raw dataset. It contains 1,048 patient cases covering 55 medical calculators, including 20 rule-based calculators and 35 numerical calculators.

For the \textbf{multi-calculator} task, we first construct cases from MedCalc-Bench by grouping entries with the same patient case and merging only their ground-truth calculator annotations, yielding 1,066 multi-calculator candidates. However, this dataset is highly imbalanced, with rule-based calculators appearing much less frequently than numerical calculators. To improve rule-based calculator coverage, we further construct cases from PMC-Patients~\citep{pmc}. Specifically, we use regular expression matching to identify clinical records that explicitly mention multiple calculator names and scores from 180,142 cases, obtaining 3,517 candidate descriptions. After filtering for cases with sufficient clinical information to compute all required calculators, we obtain 92 additional multi-calculator cases covering 33 rule-based calculators.
For all multi-calculator cases, we discard the original queries and use GPT-5.1 to generate fuzzy, goal-oriented queries from the corresponding clinical texts. This design  better simulates realistic medical calculation requests, where physicians typically describe clinical goals rather than explicitly naming all target calculators. Detailed prompts are provided in Appendix~\ref{prompt}.


For the \textbf{nested-calculator} task, we also construct data from PMC-Patients using regular expression matching to identify clinical records that mention calculators with dependency relationships. In addition to the parent calculator names, we require the presence of sub-calculator names together with their corresponding scores and supporting clinical descriptions, ensuring that both parent and sub-calculators can be derived from the clinical text. After filtering for cases with sufficient information to support the full dependency chain, we obtain 62 nested-calculator cases.

\textbf{Quality Control.}
We apply a three-stage quality-control process to ensure the reliability and realism of the constructed benchmark. 
First, to prevent answer leakage, we remove explicit mentions of calculator names and final scores from the clinical texts when such information directly reveals the target calculators or expected outputs. 
Second, three trained graduate students review all cases to verify consistency among clinical descriptions, required parameter values, and expected outputs, while ensuring that the queries are faithful to the corresponding clinical texts and clinically plausible.
Third, a clinician audits 20\% of the cases for clinical correctness and query realism. 
Unqualified cases are returned to the student reviewers with explanations for revision, and the review process is repeated iteratively until overall annotation accuracy exceeds 95\%.


\textbf{Toolkit Construction.}
We build the toolkit in two stages. First, we reuse the validated computational logic of 46 calculators from MedCalc-Bench and reorganize them under a unified configuration schema, standardizing parameter definitions and supplementing descriptions from authoritative clinical references such as the MSD Manuals\footnote{\url{https://www.msdmanuals.com/}}. 
Second, we implement 31 additional calculators to expand coverage in underrepresented areas, including neurological assessment (e.g., NIHSS), functional evaluation (e.g., 
ECOG), and psychiatric screening (e.g., PHQ-9). 
All calculators serve as callable tools within the agent framework.

\subsection{Agent Method}
To address the limitations of existing approaches in complex clinical scenarios, we propose an agent framework composed of four core modules: query rewriting, retrieval and reranking, tool selection, and tool execution. 
Detailed prompts are provided in Appendix~\ref{prompt} and a complete execution trace is provided in Appendix~\ref{Example}.

\textbf{Query Rewriting.}
In real clinical practice, physicians’ queries are often expressed in terms of task goals, such as \textit{``assessing stroke risk''} or \textit{``evaluating overall disease severity''}. Directly matching such queries with calculator descriptions often leads to inaccurate retrieval results. To better capture the underlying clinical intent, this module rewrites the original query along three medically meaningful dimensions: clinical context, key parameters, and core functionality. This multi-dimensional rewriting helps the system better align the query with relevant medical calculators and more accurately recover the physician’s intended task.

\textbf{Retrieval and Reranking.}
For each rewritten query, we perform dense retrieval~\citep{tool,retrieval} against the corresponding calculator description fields in the toolkit. Each query independently retrieves top-$K$ candidate calculators. Candidates retrieved from multiple query routes are then combined and reranked using Reciprocal Rank Fusion (RRF)~\citep{RRF}, yielding a final candidate set of $N$ calculators. The calculation formula is:
\begin{equation}
  \mathrm{RRF}(d)=\sum_{i=1}^{4}\frac{1}{k+r_i(d)}
\end{equation}

\textbf{Tool Selection.}
In multi-calculator scenarios, the number of applicable calculators for a given clinical case is unknown. Therefore, directly selecting the top-1 candidate based on similarity is insufficient. At this stage, the model is provided with the detailed descriptions and parameter requirements of all candidate calculators, enabling more fine-grained reasoning to determine which calculators should be executed.

\textbf{Tool Execution.}
Even after appropriate calculators are selected, successful execution is not guaranteed. Some complex calculators depend on sub-calculator outputs for computation. Therefore, the system first identifies dependencies, determining whether the target calculator requires inputs that cannot be directly extracted from the patient case but must be obtained by executing other calculators. If dependencies exist, relevant sub-calculators are executed first, and their outputs are used as inputs for the parent calculator.



After determining the execution order, the extracted parameters are validated through a pipeline: Completeness Check $\rightarrow$ Type Check $\rightarrow$ Enum Check $\rightarrow$ Unit Check. The system then performs an evidence audit to ensure consistency with the original clinical text and medical logic.
\begin{itemize}[leftmargin=*, itemsep=0pt, topsep=4pt, parsep=0pt, partopsep=0pt]
    \item Completeness Check: whether all parameters required for the calculation are present in the returned results;
    \item Type Check: whether the extracted values match the expected parameter types;
    \item Enum Check: whether the values of enumerated variables belong to the allowed options;
    \item Unit Check: whether the units of numerical values are consistent with calculator requirements.
\end{itemize}
Overall, if an error is detected at any stage, the system returns the error information and backtracks to the parameter extraction stage for correction. Only after all checks are satisfied will the corresponding calculators be executed to produce the final result. This mechanism helps prevent error propagation caused by incorrect parameters during nested calculator execution.

\section{Experiment}

In this section, we first conduct extensive experiments to evaluate the effectiveness of our framework. Next, we provide a detailed analysis to offer deeper insights into our framework.

\subsection{Evaluation Setup}

\textbf{Baselines.}
To evaluate the effectiveness of our framework, we compare it with three representative agent methods:
(1) MeNTi~\citep{menti}, which uses a meta-tool mechanism for tool selection and invocation;
(2) MedRaC~\citep{medrac}, which integrates retrieval-augmented generation with LLM-based code generation for medical calculation; and (3) ReAct~\citep{react}, a reasoning-and-acting framework that performs iterative step-by-step tool use.
For each method, we evaluate six LLMs to examine the interaction between model capability and framework design, including four open-source general LLMs (Qwen3-235B-A22B~\citep{qwen3}, gpt-oss-120b~\citep{gptoss}, Llama-3.3-70B-Instruct~\citep{llama3}, and DeepSeek-V3.1~\citep{deepseek}), one closed-source LLM (GPT-5-mini), and one medical-specialized LLM (MedResearcher-R1-32B~\citep{medresearcher}).



\textbf{Metrics.}
We evaluate systems across three aspects: tool selection, parameter extraction, and final score computation. We use R-F1 to measure the model's ability to select correct calculators, Extraction Accuracy to evaluate whether required parameters are correctly extracted, and Score Accuracy to assess whether each selected calculator's final score is correct. For nested-calculator tasks, we further introduce two dependency-related metrics: Dependency Detection Accuracy, which measures whether the model correctly identifies dependencies between the target calculator and its sub-calculators, and Sub-calculator Score Accuracy, which evaluates whether intermediate results from dependent sub-calculator tools are correct.

\textbf{Implementations.}
Our system is deployed through an OpenAI-compatible API interface. 
For closed-source GPT model, we use the default settings. 
For all other open-source models, we set the temperature to 0.01 to encourage stable and deterministic outputs. 
In the semantic retrieval stage, we use m3e-base~\citep{m3} as the embedding model and perform vector retrieval over the calculator toolkit using normalized cosine similarity on GPU. 
For each retrieval route, we retrieve $k_{\text{route}} = 16$ candidate tools. 
After reranking, the final number of candidates provided to the LLM selector is set to $K_{\text{cand}} = 32$. 
During tool selection, the LLM is free to determine how many tools to select from the candidate list, and we do not impose an additional upper bound on the number of selected tools. 
All experiments are conducted on a cluster with four NVIDIA A800 80GB GPUs.

\begin{table*}[t]
\centering
\tiny
\setlength{\tabcolsep}{5.5pt}
\resizebox{\textwidth}{!}{%
\begin{tabular}{llcccccc}
\toprule
\multirow{2}{*}{\textbf{Method}} & \multirow{2}{*}{\textbf{Model}} 
& \multicolumn{3}{c}{\textbf{Single-Calculator}} 
& \multicolumn{3}{c}{\textbf{Multi-Calculator}} \\
\cmidrule(lr){3-5} \cmidrule(lr){6-8}
& & \textbf{R-F1} & \textbf{Extract} & \textbf{Score} & \textbf{R-F1} & \textbf{Extract} & \textbf{Score} \\
\midrule
\multirow{6}{*}{MeNTi~\citep{menti}}
& gpt-oss-120b      & 86.11 & 37.56 & 58.40 & 29.06 & 15.12 & 15.61 \\
& Qwen3-235B-A22B   & 82.91 & 43.50 & 57.92 & 31.95 & 19.26 & 17.47 \\
& Llama-3.3-70B-Instruct & 84.53 & 38.35 & 58.49 & 33.83 & 21.11 & 26.68 \\
& GPT-5-mini        & 82.35 & 45.66 & 56.11 & 32.38 & 24.62 & 28.28 \\
& DeepSeek-V3.1     & 73.56 & 39.09 & 49.71 & 34.99 & 23.17 & 21.11 \\
& MedResearcher-R1-32B & 85.51 & 41.25 & 55.73 & 29.55 & 17.73 & 15.79 \\
\midrule
\multirow{6}{*}{MedRaC~\citep{medrac}}
& gpt-oss-120b      & 86.57 & 43.92 & 75.38 & 20.38 & 5.29  & 13.71 \\
& Qwen3-235B-A22B   & 86.57 & 42.35 & 73.85 & 18.10 & 5.53  & 11.26 \\
& Llama-3.3-70B-Instruct & 86.57 & 34.05 & 69.37 & 20.72 & 6.22 & 14.49 \\
& GPT-5-mini        & 86.57 & 44.70 & 76.72 & 18.00 & 6.97  & 12.30 \\
& DeepSeek-V3.1     & 86.57 & 39.05 & 69.08 & 21.12 & 6.86  & 12.89 \\
& MedResearcher-R1-32B & 86.57 & 40.42 & 64.50 & 20.38 & 6.33  & 12.78 \\
\midrule
\multirow{6}{*}{ReAct~\citep{react}}
& gpt-oss-120b      & 79.50 & 48.01 & 55.06 & 54.32 & 34.80 & 37.57 \\
& Qwen3-235B-A22B   & 75.51 & 39.36 & 51.34 & 50.21 & 30.38 & 33.59 \\
& Llama-3.3-70B-Instruct & 72.28 & 23.26 & 40.84 & 47.01 & 25.21 & 27.20 \\
& GPT-5-mini        & 77.90 & 35.47 & 49.05 & 50.79 & 31.99 & 35.12 \\
& DeepSeek-V3.1     & 78.03 & 46.65 & 55.06 & 48.58 & 28.91 & 30.84 \\
& MedResearcher-R1-32B & 80.21 & 40.42 & 56.49 & 44.99 & 31.82 & 34.41 \\
\midrule
\multirow{6}{*}{Ours}
& gpt-oss-120b     & \textbf{99.62} & \underline{53.56} & \textbf{80.15} & \textbf{59.97} & \textbf{53.87} & \textbf{59.53} \\
& Qwen3-235B-A22B   & 98.12 & 52.84 & \underline{79.58} & 54.44 & \underline{45.58} & 44.37 \\
& Llama-3.3-70B-Instruct & 86.63 & 37.03 & 65.94 & 53.53 & 44.53 & 46.15\\
& GPT-5-mini        & 93.13 & 41.19 & 70.13 & \underline{57.73} & 44.70 & \underline{48.94} \\
& DeepSeek-V3.1     & \underline{98.53} & \textbf{56.19} & 76.04 & 52.63 & 39.65 & 43.18 \\
& MedResearcher-R1-32B & 98.03 & 40.76 & 77.00 & 50.33 & 40.14 & 37.50 \\
\bottomrule
\end{tabular}%
}
\caption{Main results on the single-calculator and multi-calculator tasks (\%). The best and second best results are highlighted in \textbf{bold} and \underline{underline}.}
\label{tab:main_12}

\end{table*}

\begin{table*}[t]
\centering
\tiny
\resizebox{\textwidth}{!}{
\begin{tabular}{llccccc}
\toprule
\textbf{Method}                  & \textbf{Model}             & \textbf{R-F1 }   & \textbf{Dep Detect} & \textbf{Extract} & \textbf{Overall} & \textbf{Sub Overall} \\ 
\midrule
\multirow{6}{*}{MeNTi~\citep{menti}}  
                        & gpt-oss-120b      & 36.52 & 56.45 & 61.68 & 61.29 & 54.83 \\
                        & Qwen3-235B-A22B   & 37.96 & 54.84 & 59.53 & 45.16 & 58.06 \\
                        & Llama-3.3-70B-Instruct & 21.12 & 33.9 & 15.50 & 16.12 & 11.29 \\
                        & GPT-5-mini        & 31.78 & 66.13 & 56.08 & 54.83 & 53.22 \\
                        & DeepSeek-V3.1     & 34.19 & 48.39 & 53.28 & 51.61 & 45.16 \\
                        & MedResearcher-R1-32B & 42.86 & 11.29 & 40.15 & 35.43 & 19.35 \\ 
\midrule 
\multirow{6}{*}{MedRaC~\citep{medrac}} 
                        & gpt-oss-120b      & 0.00  & 0.00  & 0.00  & 0.00  & 0.00  \\
                        & Qwen3-235B-A22B   & 0.00  & 0.00  & 0.00  & 0.00  & 0.00  \\
                        & Llama-3.3-70B-Instruct & 0.00 & 0.00 & 0.00 & 0.00 & 0.00 \\
                        & GPT-5-mini        & 0.00  & 0.00  & 0.00  & 0.00  & 0.00  \\
                        & DeepSeek-V3.1     & 0.00  & 0.00  & 0.00  & 0.00  & 0.00  \\
                        & MedResearcher-R1-32B & 0.00  & 0.00  & 0.00  & 0.00  & 0.00  \\ 
\midrule 
\multirow{6}{*}{ReAct~\citep{react}}  
                        & gpt-oss-120b      & \underline{80.62} & 83.87 & 66.42 & 11.74 & 77.41 \\
                        & Qwen3-235B-A22B   & 74.02 & 75.81 & 58.45 & 19.35 & 56.45 \\
                        & Llama-3.3-70B-Instruct & 55.76 & 71.00 & 25.19 & 11.29 & 12.90 \\
                        & GPT-5-mini        & 77.78 & \underline{90.30} & 72.01 & 25.80 & 82.30 \\
                        & DeepSeek-V3.1     & 53.23 & 51.61 & 38.11 & 20.97 & 24.19 \\
                        & MedResearcher-R1-32B & 67.20 & 67.74 & 49.95 & 24.19 & 27.42 \\ 
\midrule 
\multirow{6}{*}{Ours}   
                        & gpt-oss-120b      & \textbf{87.22} & \textbf{93.50} & \underline{88.48} & 75.80 & \underline{87.10} \\
                        & Qwen3-235B-A22B   & 78.26 & 87.10 & 81.05 & \underline{75.81} & 80.65 \\
                        & Llama-3.3-70B-Instruct & 55.03 & 58.66 & 67.21 & 71.67 & 80.33 \\
                        & GPT-5-mini        & 80.00 & \textbf{93.50} & \textbf{88.91} & \textbf{83.90} & 85.50 \\
                        & DeepSeek-V3.1     & 58.73 & 59.68 & 52.53 & 51.61 & 56.45 \\
                        & MedResearcher-R1-32B & 61.54 & 64.52 & 57.27 & 56.45 & \textbf{91.94} \\ 
\bottomrule
\end{tabular}
}
\caption{Main results on the nested-calculator tasks. ``Dep Detect'' denotes Dependency Detection Accuracy, ``Overall'' denotes Parent Calculator Score Accuracy, and ``Sub Overall'' denotes Sub-calculator Score Accuracy. }
\label{tab:main_3}

\end{table*}

\subsection{Main Results}

To evaluate the effectiveness of the proposed model, we conduct systematic comparisons with all baseline methods. 
The experimental results are presented in Table~\ref{tab:main_12} and Table~\ref{tab:main_3}.


From the table, we conclude that: 
1) \textbf{Single-Calculator task performance}.
On this task, although the query explicitly specifies the target calculator, traditional methods still struggle to reliably complete tool invocation and computation. Our method improves Tool Selection R-F1 by 13.05\% and Score Accuracy by 4.77\% over the strongest baseline, MedRaC. These results indicate that existing agent frameworks lack stability even in relatively simple medical calculation scenarios.
2) \textbf{Multi-Calculator task performance}.
In this task, the performance gap further increases as task complexity grows. This setting requires selecting multiple relevant calculators for a single-patient case. Compared with the strongest baseline, ReAct, our method improves parameter extraction accuracy by 19.07\% and Score Accuracy by 21.96\%, suggesting that in multi-calculator scenarios, the primary bottleneck shifts from tool selection to parameter extraction and execution stability.
3) \textbf{Nested-Calculator task performance}.
For the nested-calculator task, method differences become more pronounced. This setting requires identifying calculator dependencies and propagating intermediate results correctly. Our method improves Sub Overall by 32.27\% and Parent Overall by 14.51\% over the strongest baseline, MeNTi. The smaller parent-level gain suggests sub-calculator execution gains only partly translate into final parent-level outputs, as residual intermediate-result errors can propagate through the dependency chain and interact with parent-calculator execution.
4) \textbf{Failure analysis of MedRaC}.
The performance collapse of MedRaC on the multi-calculator and nested-calculator tasks stems from a common limitation. Unlike the single setting, these tasks formulate queries in terms of clinical goals rather than explicit calculator names. Under this condition, the RAG-based retrieval mechanism fails to recover the corresponding formulas, preventing the execution pipeline from being established. This issue is particularly evident in the nested setting, where MedRaC fails to retrieve any valid tools, causing all metrics to drop to 0.00\%.

\subsection{Ablation Study}
To analyze the contribution of each core component, we conduct ablation experiments by removing three key modules: (1) Query Rewriting, where multi-dimensional rewriting is disabled and retrieval is performed using the original query; (2) Retrieval and Reranking, where candidate retrieval is removed and the model selects tools directly from the full calculator set; and (3) Variant-1 in the tool execution module, where the structural validation and evidence review stages after parameter extraction are removed. We use gpt-oss-120b as the base model in this experiment and all subsequent analysis experiments. The results are presented in Table~\ref{tab:ablation_o2o_o2m} and Table~\ref{tab:ablation_nested}.

\begin{table}[t]
\centering
\scriptsize
\setlength{\tabcolsep}{2pt}
\resizebox{\columnwidth}{!}{%
\begin{tabular}{@{}lcccccc@{}}
\toprule
\multirow{2}{*}{\textbf{Methods}} & \multicolumn{3}{c}{\textbf{Single-Calculator}} & \multicolumn{3}{c}{\textbf{Multi-Calculator}} \\
\cmidrule(lr){2-4} \cmidrule(lr){5-7}
& \textbf{R-F1} & \textbf{Extract} & \textbf{Score} & \textbf{R-F1} & \textbf{Extract} & \textbf{Score} \\
\midrule
Ours           & 99.62 & 53.56 & 80.15 & 59.97 & 53.72 & 59.53 \\
w/o rewrite    & 89.70 & 41.69 & 77.67 & 57.02 & 46.52 & 49.81 \\
w/o retrieval  & 79.66 & 35.95 & 67.75 & 55.59 & 41.86 & 46.08 \\
w/o variant-1 & 93.90 & 41.86 & 75.29 & 58.98 & 51.64 & 53.36 \\
\bottomrule
\end{tabular}%
}
\caption{Ablation results on the single-calculator and multi-calculator tasks.}
\label{tab:ablation_o2o_o2m}

\end{table}

\begin{table}[t]
\centering
\scriptsize
\setlength{\tabcolsep}{2pt}
\resizebox{\columnwidth}{!}{%
\begin{tabular}{@{}lccccc@{}}
\toprule
\textbf{Methods} & \textbf{R-F1} & \textbf{Dep. Detect.} & \textbf{Extract} & \textbf{Overall} & \textbf{Sub Overall} \\
\midrule
Ours           & 87.22 & 93.50 & 88.48 & 75.80 & 87.10 \\
w/o rewrite    & 79.39 & 83.90 & 75.57 & 72.60 & 77.40 \\
w/o retrieval  & 85.04 & 87.10 & 82.13 & 72.60 & 85.50 \\
w/o variant-1 & 83.97 & 88.70 & 68.78 & 64.50 & 80.60 \\
\bottomrule
\end{tabular}%
}
\caption{Ablation results on the nested-calculator task.}
\label{tab:ablation_nested}

\end{table}

The results show that:
1) Retrieval is critical for tool selection.Removing retrieval causes the largest drop in the single-calculator task: R-F1, Extract, and Score decrease by 19.96\%, 17.61\%, and 12.40\%, respectively. This indicates that even with the target calculator explicitly specified, reliable candidate filtering remains essential for stable tool selection in a large tool space.
2) Variant-1 is crucial for multi-step execution. Removing Variant-1 most severely affects nested-calculator tasks, where Extract decreases by 19.70\%, Parent Overall by 11.3\%, and Sub Overall by 6.5\%. This suggests that Variant-1 and auditing are essential to prevent parameter errors from propagating through dependency chains and affecting final results.
3) Query rewriting improves downstream reasoning. Removing query rewriting mainly affects parameter extraction and final scoring rather than tool recovery. In the multi-calculator setting, R-F1 drops by only 2.95\%, while Extract and Score decrease by 7.20\% and 9.72\%, indicating rewriting helps reconstruct task semantics from the original query and supports more reliable parameter identification and execution.

\subsection{Hyperparameter Analysis}

The retrieval stage involves two key hyperparameters: $K_{\text{route}}$ and $K_{\text{cand}}$. 
$K_{\text{route}}$ denotes the retrieval depth of each route, namely the number of candidates recalled by each retrieval branch, while $K_{\text{cand}}$ denotes the maximum size of the final candidate pool passed to the downstream Tool Selection stage.
We perform three independent grid searches on the multi-calculator tasks with $K_{\text{cand}} \in \{8,16,32,64\}$ and $K_{\text{route}} \in \{8,16,32,77\}$. 

Firstly, we determine the optimal $K_{\text{cand}}$ by comparing the average best F1 across runs (Table~\ref{tab:kcand_analysis}). 
Although F1 increases monotonically with larger candidate budgets, the marginal gain quickly diminishes: when $K_{\text{cand}}$ increases from 32 to 64, the average F1 only improves from 64.57\% to 66.73\% (+1.16\%). Meanwhile, the prompt length in the tool selection stage grows almost proportionally with the number of candidate tool descriptions, making the token consumption with $K_{\text{cand}}=64$ nearly twice that of $K_{\text{cand}}=32$. Considering the minimal performance gain and substantially higher inference cost, we adopt $K_{\text{cand}}=32$.

With $K_{\text{cand}}$ fixed, we further analyze the effect of the retrieval depth $K_{\text{route}}$ (Table~\ref{tab:kroute_analysis}). The best performance occurs at $K_{\text{route}}=16$, while larger values bring no additional benefit and slightly degrade performance.  Overall, $K_{\text{cand}}=32$ and $K_{\text{route}}=16$ provide a good balance between candidate coverage and quality.


\begin{table}[t]
\centering
\footnotesize
\resizebox{\linewidth}{!}{%
\begin{tabular}{
p{0.6cm}
>{\centering\arraybackslash}p{1cm}
>{\centering\arraybackslash}p{1cm}
>{\centering\arraybackslash}p{1cm}
>{\centering\arraybackslash}p{1cm}
>{\centering\arraybackslash}p{1cm}
}
\toprule
\textbf{$K_{\text{cand}}$} & \textbf{run1} & \textbf{run2} & \textbf{run3} & \textbf{avg} & \textbf{$\Delta$avg} \\
\midrule
8  & 59.03 & 59.13 & 59.03 & 59.06 & $-6.67$ \\
16 & 62.77 & 62.46 & 62.37 & 62.53 & $-3.20$ \\
32 & 64.66 & 64.63 & 64.43 & 64.57 & $-1.16$ \\
64$^{\dagger}$ & 65.95 & 65.73 & 65.53 & 65.73 & -- \\
\bottomrule
\end{tabular}%
}
\vspace{-0.2cm}
\caption{Validation F1 under different $K_{\text{cand}}$ values. $\Delta$avg denotes the difference from the $K_{\text{cand}}=64$ setting. $^{\dagger}$ Near-unconstrained candidate retention setting.}
\label{tab:kcand_analysis}

\end{table}

\begin{table}[t]
\centering
\footnotesize
\resizebox{\linewidth}{!}{%
\begin{tabular}{
p{1.1cm}
>{\centering\arraybackslash}p{1.8cm}
>{\centering\arraybackslash}p{1.8cm}
>{\centering\arraybackslash}p{1.8cm}
}
\toprule
\textbf{$K_{\text{route}}$} & \textbf{Precision} & \textbf{Recall} & \textbf{F1} \\
\midrule
8  & 62.52 & 66.61 & 64.50 \\
16 & 62.46 & 66.93 & 64.62 \\
32 & 62.17 & 66.43 & 64.23 \\
77 & 62.31 & 67.02 & 64.62 \\
\bottomrule
\end{tabular}%
}
\vspace{-0.2cm}
\caption{Effect of $K_{\text{route}}$ when $K_{\text{cand}}=32$.}
\label{tab:kroute_analysis}

\end{table}

\section{Conclusion}
In this work, we introduce MedCalc-Pro, a benchmark consisting of 2,268 cases and 77 calculators, incorporating three task types: single-calculator, multi-calculator, and nested-calculator calculation, which better reflect real-world clinical scenarios. To address the challenges posed by these tasks, we propose a more generalized framework that integrates dependency identification and multi-tool nested invocation. Extensive experiments demonstrate that our framework outperforms existing approaches across all task types, achieving the best performance in each case. 
We hope that our framework will help facilitate the application of LLMs in the medical field, advancing clinical decision support systems.

\section*{Limitations and Future Work}
Although the proposed framework substantially improves LLM performance on medical calculator tasks through dependency identification, nested tool execution, and structured  validation, several limitations still remain. First, the data for nested-calculator tasks is still relatively limited. Although we intentionally expanded the coverage of rule-based calculators and specifically constructed nested-calculator cases, the number of nested cases remains much smaller than that of the other two settings.

Second, the current benchmark is still mainly based on relatively clean clinical text and does not sufficiently capture the noise commonly encountered in real medical settings. Examples include OCR recognition errors, non-standard abbreviations, spelling mistakes, formatting irregularities, and cross-source document concatenation, all of which are common in real electronic medical records and scanned documents. 

Third, the current benchmark mainly adopts a single-turn query--response setting.  In real clinical scenarios, however, the input is often incomplete, and physicians or systems may need to actively ask follow-up questions, conduct multi-turn clarification, and confirm missing parameters before the calculation can be completed. As a result, the present benchmark does not assess the model's ability to proactively interact under incomplete information, nor does it capture performance in collaborative multi-turn clinical calculation.

Overall, these limitations suggest that there is still a gap between the current benchmark setting and realistic deployment conditions. Future work may therefore focus on expanding nested cases, incorporating noisier clinical text, and moving from single-turn evaluation toward more interactive multi-turn clinical calculation settings.

\section*{Ethics Consideration}

Because the medical domain is inherently high-stakes, any errors in tool selection, parameter extraction, or calculator execution may lead to incorrect clinical assessments if such systems are used without appropriate oversight. Therefore, our proposed benchmark and agent framework are intended for research purposes only and should not be used as a standalone basis for real-world diagnosis, prognosis, or treatment decisions.

Our benchmark is constructed from real-world clinical cases collected from existing benchmark resources and PMC-patient open-access dataset. The benchmark is intended for methodological evaluation rather than patient-level decision support. Although we do not intentionally include personally identifiable information, even de-identified or publicly available clinical text may still raise privacy and data governance concerns. We therefore emphasize that future extensions of this work should continue to follow applicable data protection standards and institutional review requirements.

Another ethical concern is that improved medical calculator agents may be over-interpreted as reliable clinical assistants. Although our framework improves performance in complex settings, the results also show that substantial errors remain, especially in parameter extraction and nested execution scenarios. Accordingly, these systems should be viewed as assistive research prototypes rather than clinically deployable tools, and human oversight by qualified medical professionals remains essential.


\bibliography{custom}

\appendix

\section{Appendix}

\subsection{Error Type Experiments}

To further analyze the remaining failure sources of the full framework, we manually examined 50 error cases and categorized the errors into six types. Table~\ref{tab:error_analysis} presents the distribution of the specific error types. 
Two clear phenomena emerge from the analysis. (1) Parameter-related errors constitute the majority of the remaining failures. In particular, Parameter Value Error and Enum Mapping Error account for a large proportion of the residual cases, indicating that parameter normalization remains a critical step in the pipeline. This issue becomes particularly influential in nested scenarios, where incorrect parameters produced at the sub-calculator stage may propagate along dependency chains and ultimately affect the final parent-calculator calculation.
(2) Retrieval and selection errors mainly arise in multi-tool scenarios, where a single clinical case may involve multiple relevant calculators. Errors such as Retrieval Miss and Selection Error indicate that identifying the complete set of applicable tools from a single clinical case remains challenging due to the complexity of multi-tool reasoning.

\begin{table}[ht]
\centering
\setlength{\tabcolsep}{3pt}
\resizebox{\columnwidth}{!}{%
\begin{tabular}{lccc}
\toprule
\textbf{Error Type} & \textbf{single-calculator} & \textbf{multi-calculator} & \textbf{nested-calculator} \\
\midrule
Retrieval Miss & 7 & 321 & 2 \\
Selection Error & 0 & 460 & 2 \\
Over Selection & 0 & 102 & 8 \\
Param Value Error & 468 & 116 & 27 \\
Unit Conversion Error & 75 & 55 & 0 \\
Enum Mapping Error & 386 & 66 & 23 \\
\bottomrule
\end{tabular}%
}
\caption{Error type distribution across three datasets.}
\label{tab:error_analysis}

\end{table}

\begin{table}[ht]
\centering
\small
\resizebox{\linewidth}{!}{
\begin{tabular}{
p{1.5cm}
>{\centering\arraybackslash}p{1.2cm}
>{\centering\arraybackslash}p{1.5cm}
>{\centering\arraybackslash}p{1.4cm}
}
\toprule
\textbf{Difficulty} & \textbf{Tools} & \textbf{Extract} & \textbf{Score} \\
\midrule
Easy   & 2750 & 65.22 & 68.91 \\
Medium & 817  & 47.52 & 53.98 \\
Hard   & 299  & 36.32 & 42.81 \\
\bottomrule
\end{tabular}
}
\caption{Tool-level performance by difficulty.
Difficulty levels are defined by the number of required parameters: 
Easy ($\leq$3), Medium (4--7), and Hard ($>$7).}
\label{tab:difficulty_analysis}

\end{table}

\subsection{Tool-Level Analysis by Difficulty.}
To better understand how calculator complexity affects system performance, we conduct a tool-level complexity analysis by grouping calculators according to the number of required input parameters.  The corresponding tool-level results are reported in Table~\ref{tab:difficulty_analysis}.
The results show a clear performance decline as tool complexity increases. For Easy tools, Extract and Score reach 65.22\% and 68.91\%, respectively. When the difficulty increases to Medium, these metrics drop by 17.70\% and 14.93\%, and further decrease by 28.90\% and 26.10\% for Hard tools relative to the Easy group.
This trend indicates that as the number of required parameters grows, the model faces increasing difficulty in accurate parameter extraction, and extraction errors are more likely to propagate through downstream computation, leading to lower final score accuracy. These findings highlight tool complexity as a key factor affecting system performance and further emphasize the importance of structured validation and auditing mechanisms for stabilizing execution in complex tools.

\subsection{Cost-Efficiency Analysis}
To further analyze the computational cost of different methods, we conduct a component-level efficiency analysis. Specifically, we randomly sample 100 cases from the single-calculator dataset and 100 cases from the multi-calculator dataset, and include all 62 cases from the nested-calculator dataset. For each case, we record the end-to-end running time as well as the total number of input and output tokens consumed during inference. The averaged results are reported in Table~\ref{tab:efficiency_comparison}.

\begin{table}[t]
\centering
\small
\resizebox{\columnwidth}{!}{
\begin{tabular}{ccccc}
\toprule
\textbf{Method} & \textbf{Avg Time} & \textbf{Score Acc.} & \textbf{Avg Input Tokens} & \textbf{Avg Output Tokens} \\
\midrule
MeNTi & 38.76 & 30.86 & 8,940 & 4,439 \\
MedRaC & 13.53 & 24.77 & 3,149 & 1,711 \\
ReAct & 53.29 & 31.31 & 59,136 & 3,382 \\
Ours & 37.14 & 63.06 & 22,946 & 4,857 \\
\bottomrule
\end{tabular}
}
\caption{Comparison of performance and inference cost across methods. Avg Time is measured in seconds, Score Acc. is reported in percentage, and Avg Input/Output Tokens denote the average number of input/output tokens.}
\label{tab:efficiency_comparison}

\end{table}

\section{Prompt Design for Key Components}
\label{prompt}
This section presents the detailed prompts used in several key components of our framework, including query rewriting, tool selection, dependency identification, and fuzzy query generation during data construction. These prompts are provided to enhance transparency and facilitate reproducibility of our system.

\begin{figure*}[t]
\begin{tcolorbox}[
    colframe=gray,
    colback=gray!5!white,
    coltitle=white,
    coltext=black,
    fonttitle=\bfseries,
    title=B.1 Prompt: Query Rewriting,
    boxrule=1pt,
    arc=2mm,
    width=\linewidth,
    left=7pt,
    right=7pt,
    top=5pt,
    bottom=5pt
]
\fontsize{8.5pt}{10pt}\selectfont

You are assisting a retrieval system for medical scales and clinical calculators.

\medskip
\textbf{Task:} \\
Based on the [Physician Query] and [Patient Case], generate exactly 3 search phrases for semantic retrieval, each corresponding to one of three index dimensions: \\
- Core Function \\
- Clinical Context \\
- Key Input Variables. \\
Goal: The phrases must enable retrieval of the correct scale/calculator.

\medskip
\textbf{Input:}

\textbf{Physician Query:} \{INSERT\_QUERY\_HERE\}
\textbf{Patient Case:} \{INSERT\_CASE\_HERE\}

\medskip
\textbf{Index Dimensions:}

\textbf{Core clinical function} (what clinical state, risk, severity, or outcome is assessed)\\
\textbf{Target population and clinical context} (which patient group, care setting, or stage)\\
\textbf{Key input variables} (core variables, risk factors, symptom items, or formula inputs typically required by this class of tool)\\

\medskip
\textbf{ Mandatory Rules }

\textbf{[Master Principle: Intent Lock (Focus Lock)]} \\
A) First extract the single assessment target from the physician query (e.g., pneumonia severity / AF stroke risk / VTE risk stratification / GFR estimation / upper GI bleeding triage / depression severity / pre-operative cardiac risk / MAP calculation). \\
B) All three phrases must center on this target. Other diagnoses, tests, or scales in the case that do not change "which tool should be used" must \textbf{NOT} be included.

\medskip
\textbf{[Forbidden]} \\
1) No scale, score, or rule names in any of the three phrases (e.g., APACHE, Caprini, Wells, PERC, NIHSS, GCS, PHQ are all forbidden). \\
2) Phrase 3 (Key Variables) must \textbf{NOT} simply list prominent variables from the case; list only variables that are strongly tied to the assessment target and typically used by this class of tool.

\medskip
\textbf{[Task Type Templates — for guiding Phrase 3 content]}
\begin{itemize}
    \item T1 Risk Prediction/Stratification: Phrase 3: risk factors, comorbidities, exposures, surgical/activity status.
    \item T2 Severity/Prognosis Scoring: Phrase 3: vital signs + key lab values + mental status / organ function.
    \item T3 Symptom Scales: Phrase 3: symptom items, frequency, duration, functional impact.
    \item T4 Physiological/Lab Calculators: Phrase 3: formula input variables (e.g., SBP/DBP, serum glucose/sodium, creatinine/age/sex, last menstrual period).
    \item T5 Diagnostic Probability/Exclusion Rules: Phrase 3: key discriminating features + clinical signs/lab results.
\end{itemize}

\medskip
\textbf{[Phrase Generation Requirements]} \\
Output \textbf{ONLY} a JSON array of exactly 3 strings, in the following fixed order:

\begin{itemize}
    \item \textbf{(1) Core Function Query:} \\
        Describe only the assessment target (risk/severity/probability/calculated quantity). \\
        Include common abbreviations/synonyms (e.g., CAP/pneumonia; GFR/glomerular filtration rate; VTE/DVT/PE), but \textbf{NO scale names}. \\
        Do not include patient background or specific variables.
    \item \textbf{(2) Clinical Context Query:} \\
        Describe the population and care stage/setting (e.g., emergency / ICU / pre-operative / post-operative / inpatient / outpatient; adult / pediatric / pregnant). \\
        No specific variables, no score values, no scale names. \\
        Include only context strongly relevant to tool selection.
    \item \textbf{(3) Key Variables Query:} \\
        List 5-10 variable types/risk factors/symptom items/formula inputs, comma-separated. \\
        Prioritize variables that appear in the case AND are strongly relevant to the assessment target; standard tool variables may also be included. \\
        Do \textbf{NOT} list variables or imaging findings irrelevant to the assessment target.
\end{itemize}

\medskip
\textbf{Output format:} \\
A single JSON array only, in the following format:
\begin{verbatim}
["Core Function Query here", "Clinical Context Query here", "Key Variables Query here"]
\end{verbatim}
\end{tcolorbox}
\end{figure*}

\begin{figure*}[t]
\begin{tcolorbox}[
    colframe=gray,
    colback=gray!5!white,
    coltitle=white,
    coltext=black,
    fonttitle=\bfseries,
    title=B.2 Prompt : Tool Selection,
    boxrule=1pt,
    arc=2mm,
    width=\linewidth,
    left=7pt,
    right=7pt,
    top=5pt,
    bottom=5pt
]
\fontsize{8.5pt}{10pt}\selectfont

You are a rigorous Clinical Decision Support System (CDSS) tool selector.

\medskip
\textbf{Task:} \\
Based on the physician's query intent and patient record, select the most appropriate medical scale(s)/calculator(s) from the [Candidate Tool List].

\medskip
\textbf{[Candidate Tool List]} \\
(selection must come exclusively from this list; copy strings verbatim) \\
{INSERT\_TOOL\_LIST\_HERE}

\medskip
\textbf{[Physician Query]} \\
{INSERT\_QUERY\_HERE}

\medskip
\textbf{[Patient Case]} \\
{INSERT\_CASE\_HERE}

\medskip
{INSERT\_MAX\_TOOLS\_CONSTRAINT\_HERE}

\medskip
\textbf{Selection Rules (mandatory):}
\begin{enumerate}
    \item \textbf{[Highest Priority — Exact Name Match]} If the physician query explicitly names a specific tool or scale, you MUST identify the tool whose name exactly corresponds in the candidate list. Do NOT substitute a similarly named but functionally different tool. 
   
    \item \textbf{[Computability Check]} For each candidate tool, check whether its Required Inputs can be reliably supported by the patient case text (including summary, lab results, physical exam, and explicitly negated findings).
    \item \textbf{[Tool Selection]} Output only one tool for single-tool queries. For multi-tool queries, only select tools that are necessary and applicable.
    \item \textbf{[Nested Scale Detection]} If a selected tool depends on another tool's output (child tool), declare the dependency in the "dependencies" field.
\end{enumerate}

\medskip
\textbf{Output Requirements (strict JSON; no additional text):} \\
- Output a single JSON object. No Markdown, no explanatory prefix or suffix. \\
- "selected\_tools" must be a string array (may be empty); contains only parent tools. \\
- "dependencies" must be an object: \{ "parent\_tool\_id": \{ "param": "child\_tool\_id" \} \}; use \{\} if no dependencies. \\
- "analysis" should be brief (less than 80 words); do not enumerate every candidate.

\medskip
\textbf{Output JSON structure:}
\begin{verbatim}
{
    "analysis": "brief rationale here...",
    "selected_tools": [
        "Tool ID A",
        "Tool ID B"
    ],
    "dependencies": {
        "Tool ID A": { 
        "param_name_in_A": 
        "sub_tool_id" 
        }
    }
}
\end{verbatim}
\end{tcolorbox}
\end{figure*}

\begin{figure*}[t]
\begin{tcolorbox}[
    colframe=gray,
    colback=gray!5!white,
    coltitle=white,
    coltext=black,
    fonttitle=\bfseries,
    title=B.3 Prompt: Dependency
Identification,
    boxrule=1pt,
    arc=2mm,
    width=\linewidth,
    left=7pt,
    right=7pt,
    top=5pt,
    bottom=5pt
]
\fontsize{8.5pt}{10pt}\selectfont

You are a strict clinical tool dependency checker.

\medskip
\textbf{Task:} \\
For each selected parent tool, examine its Required Inputs parameter list. Determine whether any input parameter represents a composite score that must be COMPUTED by another medical calculator (rather than directly extracted from the patient record).

\medskip
\textbf{[Doctor Query]} \\
{INSERT\_QUERY\_HERE}

\medskip
\textbf{[Patient Case]} \\
{INSERT\_CASE\_HERE}

\medskip
\textbf{[Selected Parent Tools (with their Required Inputs)]} \\
{INSERT\_SELECTED\_TOOLS\_HERE}

\medskip
\textbf{[Candidate Tool List]} \\
{INSERT\_TOOL\_LIST\_HERE}

\medskip
\textbf{[Existing Dependencies (may be empty)]} \\
{INSERT\_EXISTING\_DEPENDENCIES\_HERE}

\medskip
\textbf{Rules:}
\begin{enumerate}
    \item Only output dependencies where parent tool is in Selected Parent Tools.
    \item Child tool does NOT need to be from Candidate Tools — it can be any medical scale/calculator that exists.
    \item Do not add child tools into selected tools; only express them in dependencies.
    \item For each selected parent tool, examine its Required Inputs. If any input parameter represents a composite score or scale total , that parameter has a dependency on the corresponding sub-tool.
    \item A parameter is a dependency if its value must be COMPUTED via another scoring system, not directly read from the patient record. Raw vitals (temperature, blood pressure, heart rate) are NOT dependencies; scale totals (GCS score, NIHSS score) ARE dependencies.
    \item If no dependency is needed, return \{\}.
    \item Keep exact tool IDs and exact parameter names.
\end{enumerate}

\medskip
\textbf{Output format (strict JSON only):}
\begin{verbatim}
{
  "analysis": "short reason",
  "dependencies": {
    "Parent Tool ID": {
      "parameter_name_in_parent": "child_tool_id"
    }
  }
}
\end{verbatim}
\end{tcolorbox}
\end{figure*}

\begin{figure*}[t]
\begin{tcolorbox}[
    colframe=gray,
    colback=gray!5!white,
    coltitle=white,
    coltext=black,
    fonttitle=\bfseries,
    title=B.4 Prompt: Fuzzy Query Generation,
    boxrule=1pt,
    arc=2mm,
    width=\linewidth,
    left=7pt,
    right=7pt,
    top=5pt,
    bottom=5pt
]
\fontsize{8.5pt}{10pt}\selectfont

You are a medical benchmark dataset curator.

\medskip
\textbf{Task:} \\
Given a patient case and the ground-truth medical scale(s) that should be applied, write one physician query that naturally leads a clinician to select the specified scale(s). Do \textbf{NOT} mention the scale name(s) or acronym(s) explicitly.

\medskip
\textbf{[Patient Case]} \\
\{INSERT\_CASE\_HERE\}

\medskip
\textbf{[Ground-Truth Scale(s)]} \\
\{INSERT\_TOOLS\_HERE\}

\medskip
\textbf{[Scale Description(s)]} \\
\{INSERT\_TOOL\_DESC\_HERE\}

\medskip
\textbf{[Key Parameters the Scale Requires]} \\
\{INSERT\_MEASUREMENTS\_HERE\}

\medskip
\textbf{Requirements:}
\begin{enumerate}
    \item \textbf{Clinical authenticity} --- The query must sound like a real clinical communication (e.g., attending instruction, referral question, or bedside decision prompt); avoid academic or test-like phrasing.
    \item \textbf{No scale name leakage} --- Do \textbf{NOT} mention the scale/tool name, acronym, or any sub-score name.
    \item \textbf{Answerability} --- The query must be answerable using the patient case and the ground-truth scale(s); do not introduce information needs absent from the case.
    \item \textbf{Language} --- Output in English.
\end{enumerate}

\medskip
\textbf{Output format:} \\
A single plain string only, with no JSON wrapping and no extra text.
\begin{verbatim}
<query>Your physician query here.</query>
\end{verbatim}
\end{tcolorbox}
\end{figure*}

\section{Inference Example }
\label{Example}
This section presents a representative clinical case to illustrate the end-to-end execution process of the proposed framework. The example demonstrates how the system handles a multi-calculator scenario, including query rewriting, multi-route retrieval, tool selection, and dependency-aware execution, providing a detailed view of each stage in the pipeline.
\begin{figure*}[t]
\begin{tcolorbox}[
    colframe=gray,
    colback=gray!5!white,
    coltitle=white,
    coltext=black,
    fonttitle=\bfseries,
    title=C.Full Execution Trace of a multi-calculator Case,
    boxrule=1pt,
    arc=2mm,
    width=\textwidth,
    left=5pt,
    right=5pt,
    top=5pt,
    bottom=5pt
]
\fontsize{8.5pt}{10pt}\selectfont

\textbf{\#\# Doctor Query:} \\
For this patient with a history of atrial fibrillation, which tools should be used to assess stroke risk, the degree of neurological deficits, and functional recovery?

\textbf{\#\# 1. Query Rewriting:} \\
The system rewrote the original query into three auxiliary retrieval queries: \\
(1) \textit{stroke risk stratification, neurological deficit severity, and functional recovery assessment} \\
(2) \textit{adult patients with atrial fibrillation presenting with acute neurological symptoms in emergency or inpatient setting} \\
(3) \textit{age, hypertension, diabetes mellitus, prior stroke/TIA, heart failure, sex, systolic blood pressure, level of consciousness, motor strength, speech impairment} \\
Together with the original query, these formed four retrieval routes.

\textbf{\#\# 2. Multi-Route Retrieval:} \\
For each route, the system retrieved a ranked list of candidate tools. \\
Route 1 top-5: NIHSS, mRS Score, ECOG Performance Status, GCS, Trauma Index \\
Route 2 top-5: PHQ-9, CURB-65, NIHSS, ABCD2, MEWS \\
Route 3 top-5: NYHA, Framingham Risk, CURB-65, APACHE II Score, PHQ-9 \\
Route 4 top-5: NIHSS, ABCD2, GCS, mRS Score, Hunt--Hess Scale \\
After fusion and reranking, the final candidate set contained 32 tools, including NIHSS, mRS Score, CHA$_2$DS$_2$-VASc, ABCD2, GCS, and Framingham Risk.

\textbf{\#\# 3. Tool Selection:} \\
The system finally selected the following tools: \\
- CHA$_2$DS$_2$-VASc \\
- NIHSS \\
- mRS Score \\

\textbf{Selection Rationale:} \\
The query asks for stroke risk, neurological deficit severity, and functional recovery. These correspond naturally to CHA$_2$DS$_2$-VASc, NIHSS, and mRS, respectively.

\textbf{\#\# 4. Parameter Extraction:} \\

\textbf{NIHSS (initial extraction):} \\
The first NIHSS extraction produced a complete parameter set, including normal consciousness, normal gaze, no visual loss, no motor drift, no sensory deficit, no aphasia, and \texttt{dysarthria = severe}. \\

\textbf{mRS Score:} \\
The system extracted: \\
- \texttt{mrs\_grade = 0} \\
Supporting evidence: \\
\textit{During the outpatient neurology evaluation, the patient continued to be free of dysarthria, dysphagia, and the stigmata of TIA or CVA.} \\

\textbf{CHA$_2$DS$_2$-VASc:} \\
The extracted parameters were: \\
- congestive heart failure = no \\
- hypertension = yes \\
- age = 73 \\
- diabetes = no \\
- stroke/TIA history = no \\
- vascular disease = no \\
- sex category = female \\

\textbf{\#\# 5. Final Calculator Execution:} \\

\textbf{NIHSS:} \\
- Final score: 1 \\
- Label: \textit{Minor Stroke} \\

\textbf{mRS Score:} \\
- Final score: 0 \\
- Label: \textit{mRS Grade 0} \\

\textbf{CHA$_2$DS$_2$-VASc:} \\
- Final score: 3 \\
- Label: \textit{High Risk} \\
- Recommendation: \textit{Oral anticoagulation (OAC) is recommended} \\

\end{tcolorbox}
\end{figure*} 
\end{document}